\documentclass[fleqn,10pt]{wlscirep}
\usepackage[utf8]{inputenc}
\usepackage[T1]{fontenc}
\title{Population-level measures of perceived food access reveal barriers beyond geographic proximity}

\author[1]{Teresa Groton}
\author[1,2,*]{Benjamin Rachunok}

\affil[1]{Operations Research Department, North Carolina State University, Raleigh, 27607, United States of America}
\affil[2]{Industrial and Systems Engineering Department, North Carolina State University, Raleigh, 27607, United States of America}

\affil[*]{barachun@ncsu.edu}

\begin{abstract}

Food access is multidimensional, but population-level measurement still relies heavily on geography because perceived dimensions of access are difficult to measure at scale. Here, we use 25,125 Google Maps reviews from 49 grocery stores in Raleigh, North Carolina, to measure five dimensions of food access: availability, accessibility, affordability, accommodation, and acceptability. We identify review topics with unsupervised topic modeling and assign them to access dimensions using zero-shot classification, with 85.4\% agreement against manual coding. The resulting store-level measures capture distinct aspects of food access and reveal barriers that geographic proximity alone does not capture. Comparisons between nearby stores in the same chain further show that identical store policies can be perceived very differently across locations, consistent with food access reflecting the fit between residents and their food environment. Perceived food access also follows systematic socioeconomic and demographic patterns that broadly parallel, but do not replicate, those observed for geographic access. These results show that online grocery reviews can provide a scalable complement to geographic measures of food access.
\end{abstract}
\begin{document}

\flushbottom
\maketitle
\thispagestyle{empty}
\section*{Significance Statement}

Food access has been understood as multidimensional for decades, yet measuring these dimensions across large populations has remained difficult. Affordability, food quality, operating hours, and other characteristics determine whether available stores meet residents’ needs, but measuring these experiences has traditionally required surveys, interviews, or store audits. We show that online grocery reviews can provide scalable measures of these perceived dimensions of food access and provide a population-level estimation of these dimensions for the first time. These measures reveal barriers that geographic proximity alone misses and show that similar stores can provide very different levels of access to the populations they serve. Perceived food access also follows many of the socioeconomic and demographic patterns observed for geographic access.     Together, these results overcome a longstanding measurement barrier, enabling multidimensional analysis of food access at population scale

\section*{Introduction}
Food access is a critical determinant of diet and diet-related health outcomes, yet measuring its full range of dimensions at population scale remains a longstanding challenge \cite{caspi_local_2012,sawyer_dynamics_2021,turner_association_2021,yamaguchi_measures_2022}. Food access is inherently multidimensional, reflecting not only the spatial distribution of food resources but also whether those resources are affordable, acceptable, and responsive to residents' needs \cite{penchansky_concept_1981,caspi_local_2012,turner_association_2021}. Food access emerges from the interaction between residents and their local food environment, requiring consideration of both the characteristics of available food resources and the needs and constraints of the populations they serve\cite{yamaguchi_measures_2022,andress_juggling_2016}.  Nevertheless, populuation-level assessments continue to rely primarily on geographic measures such as store proximity, density, and travel time because these data can be collected consistently across large populations \cite{sharkey_measuring_2009,walker_disparities_2010,widener_spatial_2018}.

Neighborhood conditions are among the strongest predictors of health and well-being, underscoring the importance of accurately characterizing the environments in which people obtain food \cite{lakhani2019repurposing}. At the same time, evidence accumulated over the past two decades suggests that geographic accessibility alone provides an incomplete description of food access. Residents' perceptions of affordability, food quality, store accommodation, and other non-spatial characteristics explain variation in food purchasing and dietary behaviors that is not captured by geographic measures alone \cite{caspi_local_2012,yamaguchi_measures_2022,turner_association_2021}. Studies of perceived food environments consistently demonstrate that residents living in similar geographic environments often report substantially different experiences of food access, reflecting differences in affordability, food quality, cultural preferences, transportation, and household constraints \cite{yamaguchi_measures_2022,andress_juggling_2016}. Consequently, these dimensions are commonly measured using household surveys, resident interviews, community focus groups, and in-store audit instruments.\cite{caspi_local_2012,yamaguchi_measures_2022,glanz_healthy_2005,andress_juggling_2016}
While these approaches have substantially improved our understanding of how residents experience their local food environment, they require extensive manual data collection, limiting their spatial coverage, temporal frequency, and scalability \cite{glanz_healthy_2005,yamaguchi_measures_2022,caspi_local_2012}.

Recent advances in digital data collection have created new opportunities to characterize human experiences at population scale \cite{lazer_computational_2009,salganik_bit_2019}. Online reviews have been used to identify foodborne illness outbreaks, predict restaurant health code violations, detect human trafficking, evaluate service quality, and characterize customer experiences across a range of retail settings \cite{siering_2021,li_detecting_2023,korfiatis_measuring_2019,sutherland_topic_2020,heng_exploring_2018}. These studies demonstrate that resident-generated reviews provide reliable signals about real-world conditions and experiences. Relevant to food access, online reviews capture millions of resident-generated descriptions of grocery shopping experiences, documenting perceptions of food affordability, product quality, customer service, store operations, and other dimensions of food access that have traditionally required direct engagement with residents \cite{glanz_healthy_2005,andress_juggling_2016,yamaguchi_measures_2022}. Unlike surveys and store audits, these data are generated continuously, span diverse geographic regions, and provide a longitudinal record of how residents experience their local food environment \cite{salganik_bit_2019,lazer_computational_2009}. Despite this potential, online grocery reviews have primarily been used to study consumer sentiment, customer satisfaction, and business performance rather than as a source of population-level measures of perceived food access \cite{srivastava_reviews_2022,shen2019using}. Grocery store reviews provide an especially compelling opportunity because they capture unsolicited descriptions of affordability, food quality, customer service, store operations, and other dimensions of food access that have traditionally required surveys, interviews, and store audits.

Here, we develop a machine learning framework that enables population-level measurement of perceived food access using online grocery reviews. We demonstrate that resident-generated reviews can be used to quantify multiple perceived dimensions of food access at large scales. Using these measures, we show that geographic accessibility alone provides an incomplete characterization of food access, that access reflects the interaction between residents and their local food environment, and that perceived food access exhibits systematic socioeconomic structure. Together, these results establish online grocery reviews as a scalable data source for measuring perceived food access open new possibilities for studying food environments at population scale. Together, these results demonstrate we have overcome a longstanding measurement barrier, enabling the multidimensional study of food access at population scale for the first time.

\section*{Online grocery reviews enable population-level measurement of perceived food access}
To enable population-level measurement of perceived food access, we developed a framework that converts Google Maps grocery store reviews into store-level measures of perceived food access. We analyze 25,125 reviews spanning 2010–2025 from 49 grocery stores in and around Raleigh, North Carolina. The framework first identifies recurring themes in review text using unsupervised topic modeling (BERTopic \cite{grootendorst_bertopic_2022}), then assigns each topic to one or more dimensions of food access using zero-shot classification \cite{sarkar_zero-shot_2023}. Because reviews often describe multiple aspects of the shopping experience, topics can map to multiple dimensions when classification confidence exceeds a predefined threshold (Methods). We combine these dimension assignments with Google star ratings to generate review-level signals, which are then aggregated into store-level scores for each dimension. Differences in review counts are accounted for using Bayesian partial pooling, and scores are standardized within each dimension across the study region, allowing them to be interpreted as standard deviations above or below the regional mean (Methods). Automated topic assignments agreed with manual labels for 85.4\% of topics.

We interpret online grocery reviews using the multidimensional framework for food access proposed by Caspi et al., building on the Theory of Access developed by Penchansky and Thomas \cite{caspi_local_2012,penchansky_concept_1981}. The framework organizes review content into five complementary dimensions—availability, accessibility, affordability, accommodation, and acceptability—that together characterize residents’ interactions with their local food environment (Table \ref{tab:food_access_dimensions}).
\begin{table}[ht]
\centering
\small
\caption{The five dimensions of food access used to interpret online grocery reviews, with representative review excerpts from the study dataset.}
\label{tab:food_access_dimensions}
\begin{tabular}{p{1.2in} p{2.2in} p{2.8in}}
\toprule
\textbf{Dimension} & \textbf{Description} & \textbf{Example Review Excerpt} \\
\midrule

Availability &
Whether desired foods are consistently stocked and available. &
\emph{``They always have stuff available in stock when you need it.''}
\\[0.5em]

Accessibility &
Ease of reaching and entering the store, including transportation and parking. &
\emph{``Parking is terrible.''}
\\[0.5em]

Affordability &
Perceived food prices and value relative to cost. &
\emph{``We go out of our way for the deals offered.''}
\\[0.5em]

Accommodation &
Extent to which the store meets residents' needs, including operating hours, payment options, and dietary requirements. &
\emph{``They decided to close before I get off work—one star if I can't shop there!''}
\\[0.5em]

Acceptability &
Food quality, cleanliness, customer service, and alignment with residents' preferences. &
\emph{``The place is a little grimy and the produce selections are subpar.''}
\\

\bottomrule
\end{tabular}
\normalsize
\end{table}
\begin{figure}[!tbh]
\centering
\includegraphics[width=\linewidth]{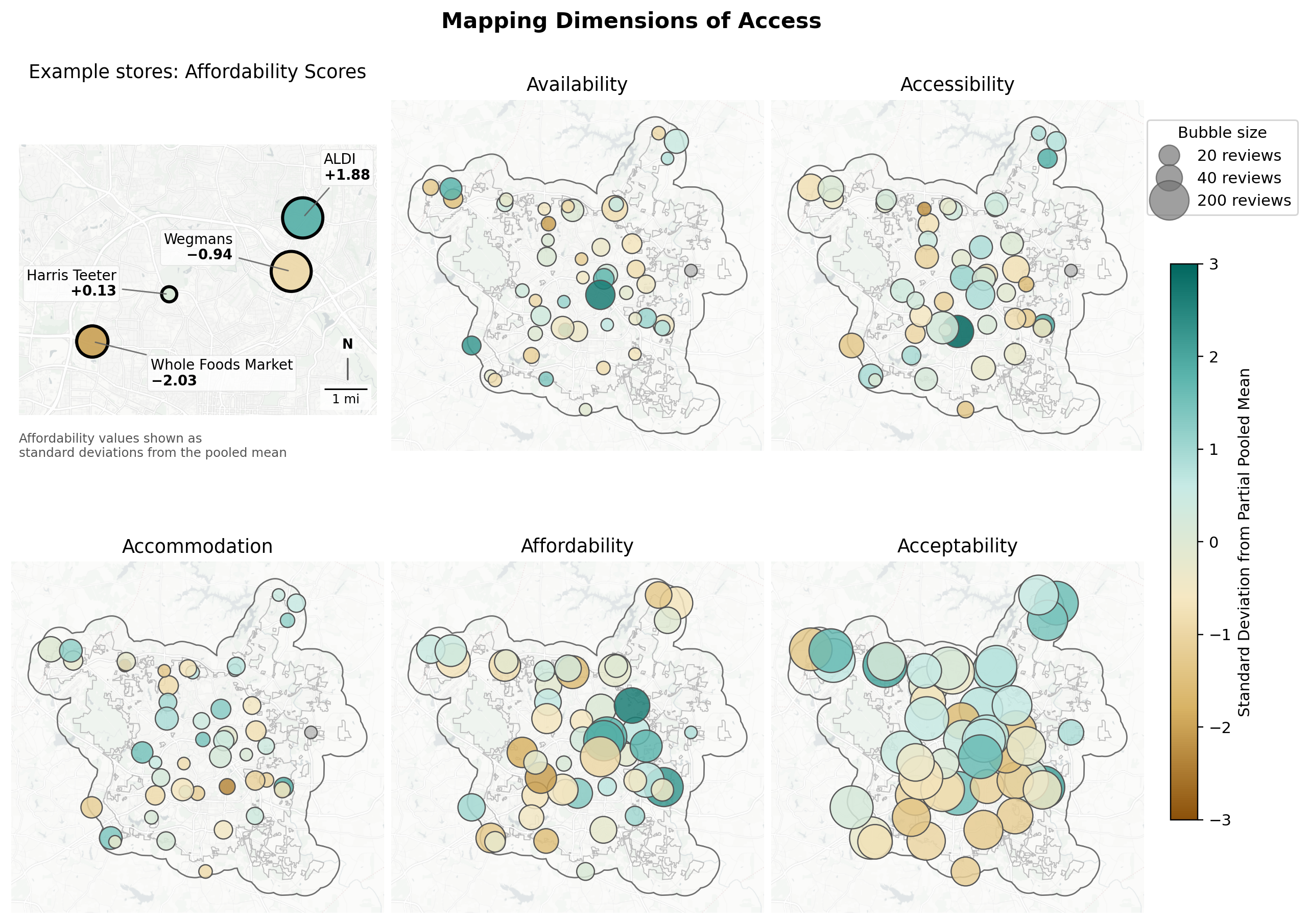}
\caption{\textbf{Online grocery reviews reveal distinct dimensions of perceived food access across Raleigh grocery stores.} a, Affordability scores for four nearby grocery stores, illustrating variation in perceived affordability among geographically proximate retailers. b–f, Store-level scores for availability, accessibility, accommodation, affordability, and acceptability across the study region. Circle size indicates the number of reviews contributing to each store–dimension score, and color indicates the standardized partially pooled score relative to the regional mean. Positive values indicate higher perceived access and negative values indicate lower perceived access within each dimension. Store scores vary substantially across dimensions, indicating that stores that perform well on one dimension do not necessarily perform well on others. }
\label{fig:mapping}
\end{figure}

Applying the framework to grocery stores in Raleigh produced store-level measures for each of the five dimensions of food access (Fig. \ref{fig:mapping}). Of the 25,125 reviews analyzed, 16,790 (66.8\%) were assigned to at least one dimension. Acceptability was the most prevalent dimension, appearing in 12,668 reviews (50.4\% of all reviews), followed by affordability (12.1\%), accessibility (4.5\%), accommodation (2.7\%), and availability (2.6\%). The prevalence of acceptability reflected frequent discussion of food quality, cleanliness, and customer service, whereas operational characteristics such as store hours and product availability appeared less frequently. Despite differences in prevalence across dimensions, their relative distribution was broadly consistent across stores and grocery chains (Supplementary Table 3), suggesting that differences in store-level scores were not driven primarily by systematic differences in the types of access discussed at different stores.

We find the resulting scores distinguish stores in ways that were consistent with well-known differences in their business models (Figure \ref{fig:mapping} a). Among the four adjacent grocery stores highlighted in the inset of Figure \ref{fig:mapping}, Aldi received the highest affordability score in the study region (+1.88 standard deviations), consistent with its emphasis on low-cost grocery retailing. Harris Teeter (a quality-focused Kroger chain common in the Southeast US) scored near the regional average, whereas Wegmans (a eastern US chain focused on customer service and high quality options) scored approximately one standard deviation below average. Whole Foods (a US based chain that markets fine natural and organic foods) received the lowest affordability score of all 49 stores (-2.03 standard deviations), reflecting widespread discussion of higher prices in customer reviews. Importantly, these differences were dimension-specific rather than indicative of overall store quality. For example, the same Aldi location that ranked highest for affordability scored only +0.22 standard deviations for accessibility, illustrating that stores exhibit distinct strengths and weaknesses across different dimensions of food access.

We find that stores exhibit distinct patterns across the five dimensions of food access rather than consistently performing well or poorly across all dimensions. For example, the Aldi location shown in Figure \ref{fig:mapping}a received the highest affordability score in the study region (+1.88 standard deviations) but scored only +0.22 standard deviations for accessibility. Conversely, the Whole Foods location received the lowest affordability score (-2.03 standard deviations) while scoring above the regional average for acceptability, illustrating that stores exhibit distinct strengths and weaknesses across different dimensions of food access. Overall, only two of the 49 stores scored at or above the regional average across all five dimensions, and no store exceeded +0.20 standard deviations above the regional mean for every dimension. Pairwise correlations between store-level dimension scores were generally weak, with all absolute correlations below 0.40 ($|r| \leq 0.39$), indicating that the five measures captured largely distinct aspects of food access. The strongest associations were between accessibility and accommodation ($r = 0.39$) and between accessibility and acceptability ($r = 0.35$).

\begin{figure}[!t]
\centering
\includegraphics[width=\linewidth]{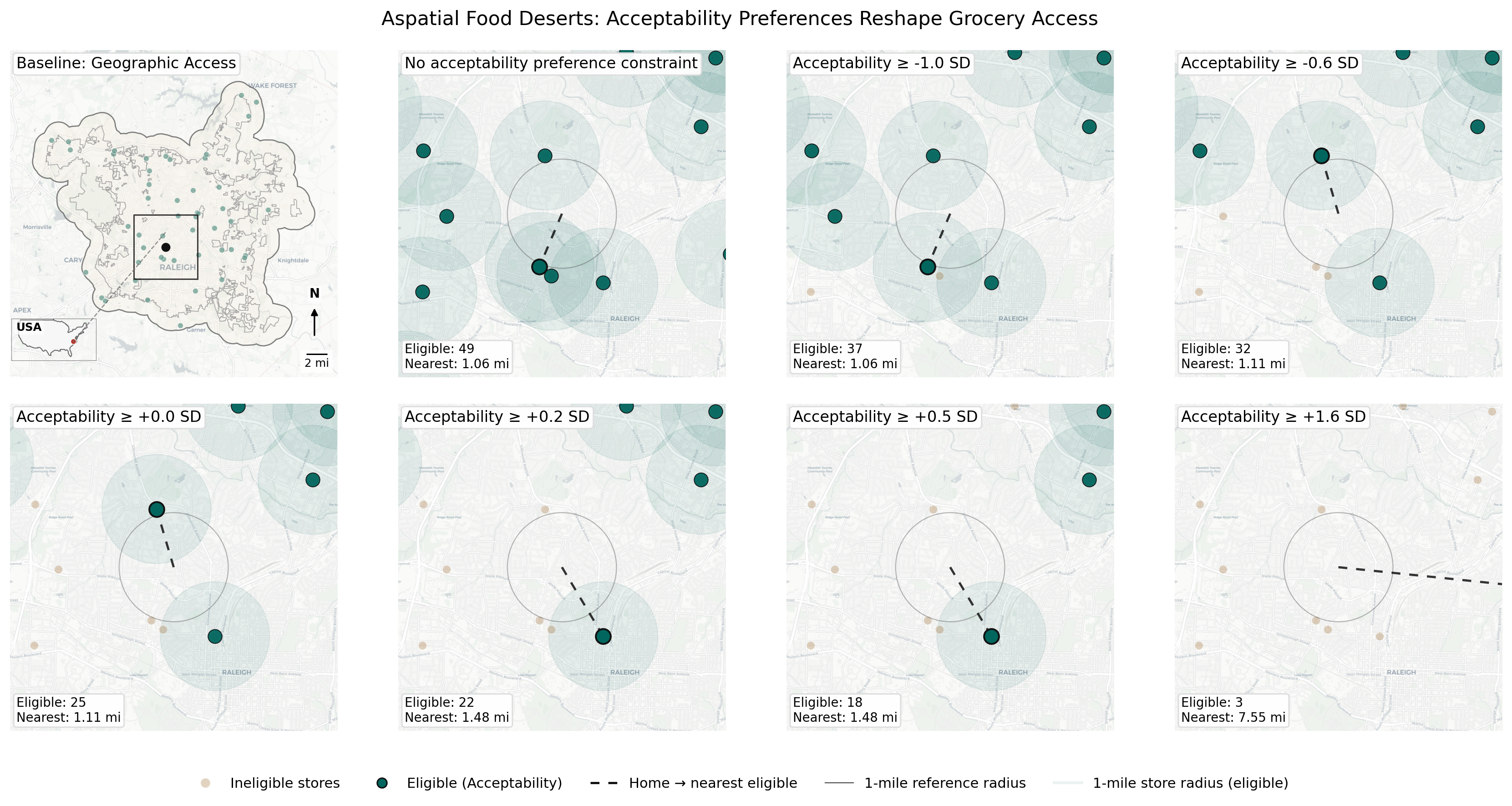}
\caption{\textbf{Perceived food access constraints reduce residents’ grocery-store opportunity sets.} a, Location of the example resident within the Raleigh study area. b–h, Grocery stores available to the example resident under progressively more restrictive acceptability requirements. Stores satisfying each minimum acceptability threshold are shown as eligible, while stores below the threshold are excluded from the resident’s opportunity set. The dashed line connects the resident to the nearest eligible store; the solid circle indicates a one-mile reference radius, and shaded circles indicate one-mile radii around eligible stores. As the minimum acceptable store score increases, fewer stores remain in the opportunity set and the distance to the nearest eligible store increases. Store acceptability scores are standardized relative to the regional mean.}
\label{fig:desert}
\end{figure}

\section*{Geographic proximity overestimates perceived food access}

We next examined how incorporating perceived food access changes conventional measures of geographic food access. Geographic measures commonly characterize access using distance to the nearest food retailer, the number or density of retailers within a specified area, or travel time to available stores \cite{sharkey_measuring_2009,walker_disparities_2010,widener_spatial_2018}. For example, the U.S. Department of Agriculture identifies low-access urban census tracts partly based on the share of residents living more than one mile from a supermarket or large grocery store \cite{economic_research_service_us_department_of_agriculture_food_nodate}. These measures characterize whether food retailers are geographically available but generally do not account for whether nearby stores meet residents’ needs and preferences. Here, we define a resident’s \textit{opportunity set} as the subset of geographically available stores that satisfy their perceived needs, preferences, and constraints. Because these factors vary across individuals, residents with identical geographic access may experience substantially different levels of food access \cite{caspi_local_2012,turner_association_2021,yamaguchi_measures_2022,sawyer_dynamics_2021}. Figure \ref{fig:desert}b–h illustrates this concept by progressively restricting the opportunity set according to increasingly stringent thresholds for perceived food access.

We find that residents’ opportunity sets contract as perceived food access is considered. Figure \ref{fig:desert} begins with the conventional geographic measure of food access, where the hypothetical resident has three grocery stores within approximately one mile of their home (Fig. \ref{fig:desert} b). Imposing a modest minimum threshold for perceived acceptability (in this example, we assume residents will not shop at stores in the bottom 15\% of acceptability scores) removes one nearby store from the resident’s opportunity set. As the required acceptability score increases, additional stores no longer satisfy the resident’s criteria. Under the strictest threshold shown (residents will only shop at stores in the top 7\% of acceptability), the nearest acceptable grocery store is 7.5 miles away.

The patterns shown in Figure \ref{fig:desert} were consistent throughout the study region and across all five dimensions of food access. Regardless of location or access dimension, increasingly stringent thresholds for perceived food access progressively reduced the set of grocery stores available to residents. Because the five dimensions capture distinct aspects of food access and are only weakly correlated, simultaneously considering multiple resident requirements further compounds this reduction. Consequently, the difference between geographic proximity and perceived food access increases as residents’ needs, preferences, and constraints become more numerous or more restrictive.

We note these results should not be interpreted as predicting where residents ultimately choose to shop. Rather, they characterize the set of stores that satisfy residents’ perceived requirements. Food store selection is inherently a multi-attribute decision, with shoppers simultaneously weighing geographic proximity alongside factors such as price, quality, convenience, and personal preferences \cite{krukowski2013there,krukowski2012qualitative}.
Consequently, these results may indicate residents may travel farther to reach stores that better satisfy their needs, or it may indicate that residents instead choose nearby stores while accepting compromises in affordability, food quality, convenience, or other dimensions of food access. The framework developed here characterizes these tradeoffs by identifying the opportunity set available to residents, rather than predicting realized shopping behavior.

\section*{Food access is jointly determined by store characteristics and resident needs}
\begin{figure}[!tbh]
\centering
\includegraphics[width=\linewidth]{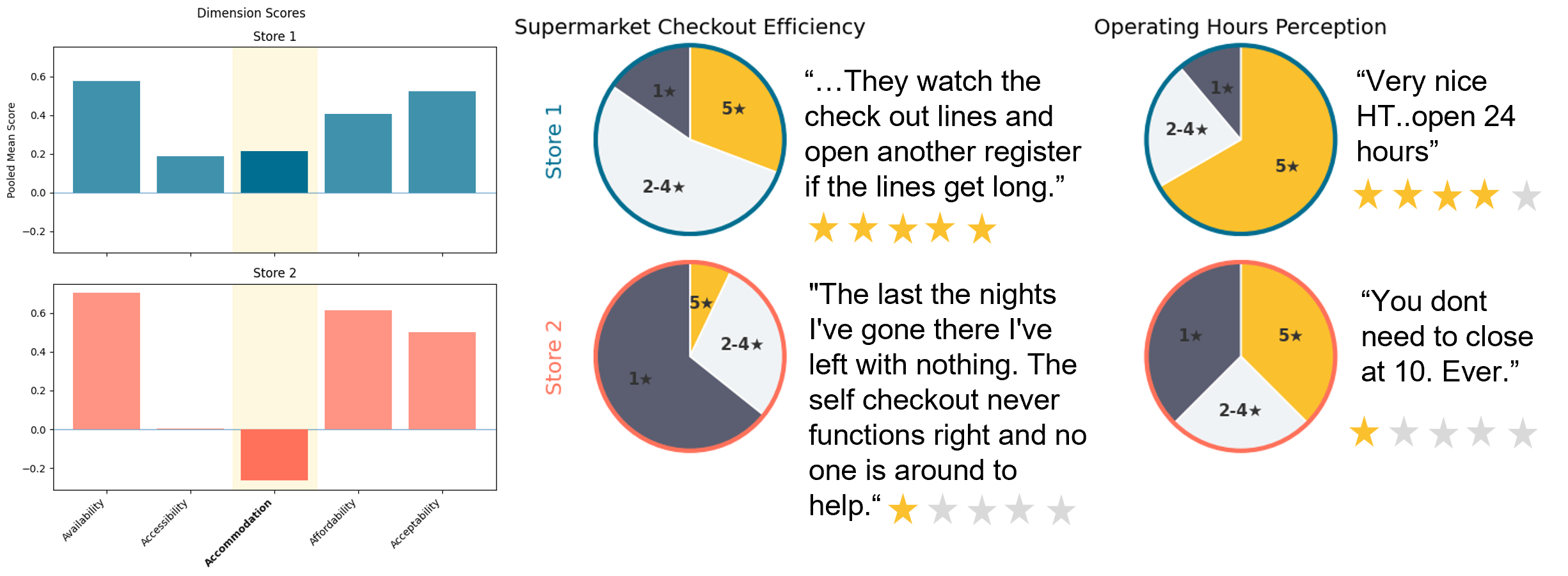}
\caption{\textbf{Nearby stores operating under similar conditions can differ substantially in perceived food access.} a, Comparison of two Harris Teeter locations less than one mile apart in western Raleigh. b, Standardized perceived food-access scores for the two stores across the five dimensions of access. The stores received similar scores across most dimensions but differed substantially in accommodation. c–d, Review content associated with accommodation reveals differences in perceptions of checkout efficiency and operating hours between the two stores. Both locations ended 24-hour operation on March 4, 2020, as part of the same chain-wide policy change. Following this change, approximately 80\% of accommodation-related reviews at the lower-scoring store discussed reduced operating hours, whereas operating hours were not mentioned in accommodation-related reviews at the higher-scoring store. This contrast illustrates how the same store characteristic can generate different perceptions of access across locations.}
\label{fig:stores}
\end{figure}

We examined pairs of nearby, similar grocery stores belonging to the same retail chain to identify differences in perceived food access that could not readily be explained by store characteristics alone. These pairs were selected to minimize differences in brand, pricing structure, product offerings, and corporate policies, allowing resident perceptions to be compared under otherwise similar conditions. Figure \ref{fig:stores} presents one such pair of Harris Teeter  stores both located in the greater western Raleigh area. The two stores received similar overall customer ratings (4.22 stars versus 4.28 stars averaged across all reviews) and comparable scores across most dimensions of food access, differing primarily in accommodation (Fig. \ref{fig:stores},a). Using the intermediate topic modeling step (Methods), we traced store-level accommodation scores back to the underlying review content, allowing the specific experiences contributing to these differences to be identified.

Reviews of the two stores revealed different perceptions of checkout efficiency despite the stores belonging to the same retail chain and operating under similar corporate policies. Differences in local management or staffing could plausibly contribute to these contrasting experiences. However, an even stronger contrast emerged for operating hours.     On March 4, 2020, Harris Teeter ended 24-hour operation across its Raleigh stores, and both locations subsequently operated under the same reduced hours \cite{prosser_harris_2020}. Following this chain-wide policy change, approximately 80\% of accommodation-related reviews at the lower-scoring store discussed the shortened operating hours, whereas the same issue was never mentioned in accommodation-related reviews at the higher-scoring store. Because both stores experienced the same policy change, these contrasting reviews provide evidence that the same operating characteristic can be perceived differently across locations. More broadly, these findings support the longstanding view that food access emerges from the interaction between residents and their food environment rather than from either in isolation \cite{penchansky_concept_1981,caspi_local_2012,sawyer_dynamics_2021,yamaguchi_measures_2022}. Identical store characteristics can therefore present different barriers depending on the needs and constraints of the residents interacting with them.

\section*{Perceived food access is structured by neighborhood socioeconomic characteristics}
We compared socioeconomic and demographic trends in perceived food access against well-established patterns observed for geographic food access. A large body of work has shown that geographic food access varies systematically across neighborhood income and racial composition \cite{morland_disparities_2007,morland_neighborhood_2002,darmon_does_2008,walker_disparities_2010,sawyer_dynamics_2021}. To evaluate whether similar patterns emerge for perceived food access, we first aggregated store-level dimension scores to census tracts using a distance-decay spatial interaction model. We then relate both perceived food access and geographic food access to neighborhood socioeconomic and demographic characteristics by estimating  weighted least-squares regression models (Methods).

We find that perceived food access varies systematically with neighborhood socioeconomic and demographic characteristics, although these relationships differ across dimensions (Table \ref{tab:regression}). Census tracts with larger Black populations were associated with lower perceived availability, accessibility, accommodation, and acceptability, but higher perceived affordability. Of these relationships, the associations with availability and affordability were statistically significant. Income was most strongly associated with acceptability, with higher-income census tracts having substantially higher perceived acceptability scores. Acceptability also showed the strongest overall socioeconomic patterning among the five dimensions ($R^2=0.415$). Together, these results show that perceived dimensions of food access are systematically patterned across neighborhoods, much like geographic access, but that the nature of these patterns depends on the dimension considered.

However, perceived food access and geographic food access were not always aligned. Consistent with prior work, census tracts with a higher percentage of Black residents and higher median household incomes were located farther from the nearest grocery store. These same socioeconomic patterns were not uniformly reflected across the perceived dimensions of food access. For example, geographic measures identified poorer access in census tracts with a higher percentage of Black residents, and incorporating perceived availability further reinforced this disadvantage. In contrast, perceived affordability was higher in these same communities. Higher-income neighborhoods exhibited a different form of decoupling: despite greater distance from grocery stores, they had substantially higher perceived acceptability. We conclude that incorporating perceived food access does not uniformly increase or decrease measured disadvantage. Rather, it reveals dimensions of access that reinforce geographic disparities, others that offset them, and communities whose food access is poorly characterized by geography alone.

\begin{table}[htp]
\centering
\caption{Weighted OLS Results: Distance and Food Access Dimensions}
\label{tab:regression}
\begin{tabular}{lcccc}
\toprule
Outcome & Percent Black (c) & Income (c) &  Percent Black $\times$ Income & $R^2$ \\
\midrule
Distance to Nearest Store (z) 
    & 4.078*** & 0.204*** & 0.680*** & 0.358 \\

Availability 
    & -0.479* & -0.016 & -0.024 & 0.063 \\

Accessibility 
    & -0.238 & 0.005 & 0.016 & 0.028 \\

Accommodation 
    & -0.149 & 0.020 & 0.174* & 0.100 \\

Affordability 
    & 1.640*** & 0.041* & 0.062 & 0.258 \\

Acceptability 
    & -0.123 & 0.099*** & 0.210** & 0.415 \\

\bottomrule
\end{tabular}

\vspace{0.5em}
\footnotesize
\textit{Notes:} Coefficients shown. Covariates are centered; median household income is expressed in \$10,000 units.
* $p<0.05$, ** $p<0.01$, *** $p<0.001$.
\end{table}

\section*{Discussion}

Our results demonstrate that online grocery reviews can provide population-level measures of perceived food access that complement conventional measures of geographic accessibility. These measures capture distinct dimensions of residents’ experiences with their food environment and reveal barriers that geographic proximity alone does not identify. Importantly, incorporating perceived food access can change how disparities across communities are characterized: perceived dimensions may reinforce geographic disadvantage in some communities while offsetting it in others. Together, these findings address a longstanding measurement challenge in food-access research by enabling multidimensional measures of perceived access to be incorporated into population-scale analyses.

Our results also highlight the importance of measuring food access as the fit between residents and their food environment. Some dimensions of access can be partially characterized from store attributes alone, such as product availability, prices, or operating hours. However, these characteristics do not establish whether a store meets residents’ needs. This distinction was particularly apparent for accommodation: two nearby stores in the same chain adopted the same change in operating hours, yet reviews at one location frequently identified the change as a barrier while reviews at the other did not. Thus, identical store characteristics can produce different levels of perceived access depending on the needs and constraints of the populations interacting with them. Measuring dimensions such as accommodation and acceptability therefore requires information about both the food environment and residents’ experiences within it.

Accordingly, the content of online reviews also provides an opportunity to refine how the less frequently measured dimensions of food access are operationalized. Although acceptability and accommodation are well established conceptually \cite{widener_spatial_2018,caspi_local_2012}, their empirical measurement often relies on supply-side characteristics. Store audits, for example, may characterize acceptability using predetermined measures of product variety and selection \cite{turner_association_2021}, while accommodation is commonly represented by store metadata such as operating hours and payment options \cite{thornton_neighbourhood-socioeconomic_2010,thatcher_retail_2017,konapur_5_2022,jiao_developing_2025}. These measures describe characteristics of the food environment but provide limited information about whether those characteristics meet residents’ needs. The matched-store comparison in Fig. \ref{fig:stores} illustrates this distinction: stores subject to the same change in operating hours generated markedly different resident responses. Supply-side measures would characterize the change identically, whereas resident-generated data reveal substantial differences in perceived accommodation. 

Open-ended review text further suggests that the empirical scope of these dimensions may be broader than commonly operationalized. Our topic model identified ten distinct themes concerning customer service and social interactions, alongside themes such as store cleanliness, as important components of acceptability. These experiences may be missed by instruments defined a priori. Similar discrepancies between objective conditions and resident perceptions have been observed elsewhere; for example, residents may avoid otherwise available foods because of perceived concerns about quality or safety \cite{konapur_5_2022}. Unsolicited, open-ended data can therefore complement structured measurement by identifying locally salient components of food access that researchers may not specify in advance. Future work applying vector embedding to grocery store reviews could be used to identify latent dimensions of access which may not be present in existing frameworks. 

Incorporating perceived food access can also change how disparities in food access are identified. Geographic and perceived access were systematically patterned across neighborhoods, but these patterns did not always align. In census tracts with a higher percentage of Black residents, for example, poorer geographic access was reinforced by lower perceived availability but offset by higher perceived affordability. Higher-income neighborhoods showed a different divergence, with greater geographic distance to grocery stores but substantially higher perceived acceptability. Thus, relying on geography alone can differentially characterize food-access disadvantage across communities, depending on which dimensions of access are considered. This may help explain why geographic measures of the food environment do not consistently correspond to residents’ experiences or dietary outcomes \cite{caspi_local_2012,walker_disparities_2010,sawyer_dynamics_2021,turner_association_2021}.

Our measures characterize constraints on residents’ food environments rather than their realized shopping behavior. Residents facing a limited set of stores that meet their needs may travel farther to reach a preferred retailer or continue shopping nearby while compromising on affordability, quality, convenience, or other dimensions of access. Prior work documents both types of trade-offs in grocery-store choice \cite{andress_juggling_2016,thatcher_retail_2017}. Distinguishing between these responses will require combining measures of perceived food access with information on realized shopping behavior, such as household travel or mobility data.

Several limitations should be considered when interpreting these measures. First, Google Maps reviewers are a self-selected sample and are not necessarily representative of all residents who interact with a store \cite{hu_can_2006}. The resulting scores should therefore be interpreted as measures of expressed perceptions rather than population-representative estimates of resident experience. We evaluated one source of potential reviewer-selection bias by comparing users who reviewed only one grocery store in our sample with those who reviewed multiple stores. Although multi-store reviewers assigned somewhat higher ratings overall, differences in dimension-specific scores were generally small, and the distribution of dimensions discussed was similar between the two groups (Supplementary Tables 4–5).

Second, our store-level measures are relative to the retail market in which they are estimated. Because scores were standardized within each dimension across stores in the study region, they indicate whether a store performs better or worse than other stores in the same market rather than providing an absolute measure of perceived access. Comparisons using these standardized scores should therefore be restricted to coherent retail markets in which stores plausibly compete for the same shoppers. Future applications could instead retain the underlying unstandardized scores to support comparisons across regions or use mobility data to define store-specific retail markets more directly.

Third, associations between neighborhood characteristics and perceived food access may reflect both resident experiences and retailer location decisions. Grocery chains select locations in response to market conditions, and stores serving different populations may differ systematically in pricing, product offerings, or other characteristics. For example, the higher perceived affordability observed near census tracts with a larger Black population could reflect both residents’ perceptions of nearby stores and the types of retailers operating in those areas. Our cross-sectional analysis cannot separate these processes. Future work incorporating changes in store openings, closures, or operating characteristics could help distinguish changes in perceived access from endogenous patterns of retailer location.

Despite these limitations, the framework provides a scalable approach for incorporating perceived dimensions into population-level measures of food access. Topic identification and dimension assignment rely on pretrained language models rather than region-specific supervised training, allowing the approach to be applied in other locations where sufficient review data are available. More broadly, resident-generated data could complement conventional measures of access in other domains where service availability alone does not establish whether resources meet residents’ needs, including healthcare and other essential services. Incorporating these demand-side experiences alongside measures of resource availability provides a path toward population-level measures of access that better reflect the interaction between people and the resources available to them.

\section*{Methods}
\begin{figure}[!h]
    \centering
    \includegraphics[width=\linewidth]{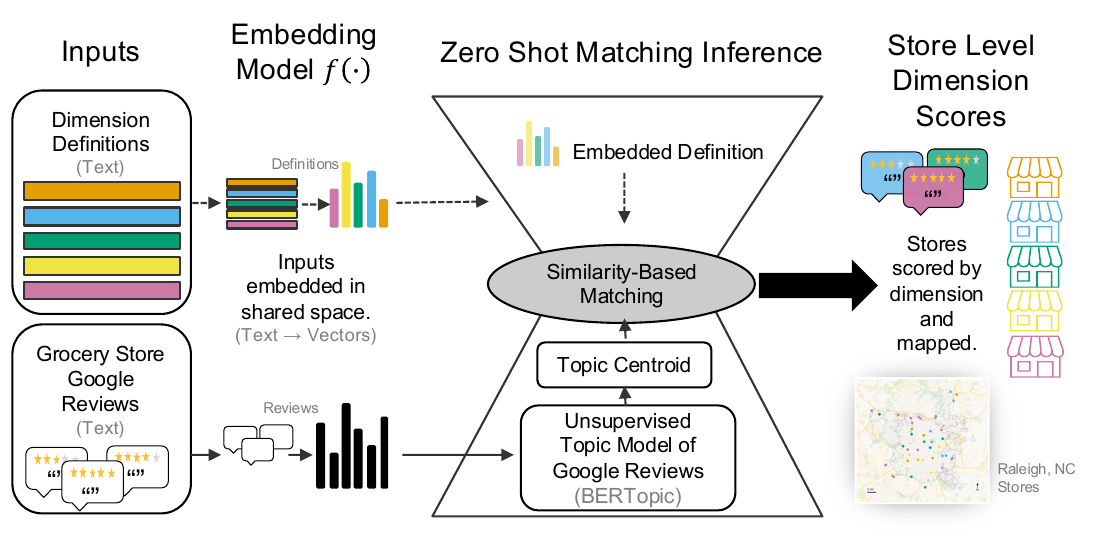}
    \caption{Overview of review-topic matching inference}
    \label{fig:Overview}
\end{figure}

\subsection*{Study area and review corpus}
We used Raleigh, North Carolina, as a case study. The study area comprised the municipal boundary and a 1-mile buffer, selected to include stores that could plausibly serve residents near the city boundary under the US Department of Agriculture's 1-mile urban food-access threshold \cite{economic_research_service_us_department_of_agriculture_food_nodate}. We included 49 nationally recognized grocery stores: 39 within the municipal boundary and 10 within the buffer. We excluded mixed-use retailers, convenience stores and other outlets for which grocery retail was not the primary business.

Google Maps reviews were collected using the Compass Google Maps Reviews Scraper through the Apify platform \cite{apify_gmaps_reviews_2026}. The source records contained review text, star rating, publication date, store location and associated metadata. We collected all reviews available at the time of extraction (June of 2025). After removal of reviews that lacked usable text or were classified as incoherent by the topic model, the analytical corpus contained 25,125 reviews published from 2010 to 2025. The median number of retained reviews per store was 430 (interquartile range, 274--669; range, 54--2,380).

We obtained census-tract population, median household income and the proportion of residents identifying as Black from the 2022 American Community Survey 5-year estimates \cite{acs_2022}.

\subsection*{Text preprocessing and topic modeling}
We converted review text to lowercase and removed punctuation, HTML tags, common English stop words and custom stop words containing grocery-chain and store-name variants. We encoded each processed review using the Sentence Transformers \texttt{all-MiniLM-L12-v2} model, which maps text to a 384-dimensional embedding space \cite{allminilm_l12_v2_2024,reimers_sentence-bert_2019,wang_minilm_2020}.

We identified recurring review themes using BERTopic \cite{grootendorst_bertopic_2022}. Review embeddings were reduced to five dimensions using UMAP with 30 neighbours, minimum distance 0.05 and cosine distance \cite{mcinnes_umap_2020}. We then clustered the reduced embeddings using HDBSCAN with a minimum cluster size of 120, minimum sample size of 15, Euclidean distance and leaf cluster selection \cite{campello_density-based_2013,mcinnes_accelerated_2017}. Reviews assigned to the HDBSCAN noise class were excluded from topic-based analysis. Topics were represented using the ten highest-weighted terms from class-based term frequency--inverse document frequency. Short descriptive titles were generated from these terms using GPT-3.5 and were used only for presentation; downstream dimension assignment used review embeddings rather than the generated titles \cite{openai_gpt35_2026}.

We compared eight candidate topic models using semantic coherence, topic diversity and manual assessment of interpretability. Topic diversity was calculated as
\begin{equation}
\mathrm{TD}=\frac{W}{K N},
\label{eq:topic_diversity}
\end{equation}
where $W$ is the number of unique words among the retained topic terms, $K$ is the number of topics and $N=10$ is the number of terms retained per topic \cite{roder_exploring_2015,dieng_topic_2020,wu_survey_2024}. The selected model produced 59 topics (coherence, 0.528; diversity, 0.709). We excluded 18 topics that were incoherent or did not contain substantive information about food access, leaving 41 topics for dimension assignment.

\subsection*{Mapping topics to dimensions of food access}
We mapped topics to the five dimensions of food access---availability, accessibility, affordability, accommodation and acceptability---using zero-shot matching inference \cite{sarkar_zero-shot_2023}. For each dimension, we constructed a prototype definition written in the style of a grocery review (Table \ref{tab:definitions}). Dimension definitions and reviews were embedded with the same \texttt{all-MiniLM-L12-v2} model and normalized to unit length. For topic $k$, we calculated the centroid $\mathbf{c}_k$ as the coordinate-wise mean of the normalized embeddings of reviews assigned to that topic and renormalized the centroid. Similarity between dimension $j$ and topic $k$ was the cosine similarity $s_{jk}$ between the normalized dimension embedding and $\mathbf{c}_k$.

\begin{table*}[t]
\centering
\caption{Food-access dimension definitions used for zero-shot embedding and topic--dimension matching. Definitions reproduce the text supplied to the embedding model, with minor typographical and punctuation changes for ease of reading. Definitions used for embedding also had preprocessing performed on them.}
\label{tab:definitions}
\small
\setlength{\tabcolsep}{6pt}
\renewcommand{\arraystretch}{1.15}
\begin{tabular}{@{}p{0.16\textwidth}p{0.80\textwidth}@{}}
\toprule
Dimension & Definition \\
\midrule
Availability & \textit{It makes a difference that there is even a grocery store in the neighborhood, and that it carries enough healthy food for the people who live nearby. People notice if there are plenty of stores around that match the size of the community, and if those stores keep a good supply of fresh food in stock. It matters when the shelves are consistently stocked with healthy staples instead of being empty or limited. }\\
\addlinespace
Accessibility & \textit{It is easy or hard to get to the store depending on distance, travel time, and whether you have good transportation options. Some people feel the store is close and convenient, while others see it as too far or difficult to reach without a car. Barriers like poor transit, unsafe roads, parking, or mobility issues make a store less accessible.} \\
\addlinespace
Accommodation & \textit{The store makes it easier or harder for people to shop depending on its hours, layout, and organization. Things like payment options, whether they take SNAP or WIC, or if they offer curbside pickup matter for how usable the store feels. People care when a store provides options for dietary restrictions, like gluten-free options.} \\
\addlinespace
Affordability & \textit{Food here can feel affordable or expensive depending on the prices and whether people think it is worth the cost. Shoppers compare what they pay here to what they get, and sometimes feel the prices are too high or overpriced. But people like the deals and discounts this store offers. The cost of groceries makes it easier or harder for families to keep buying nutritious food regularly.} \\
\addlinespace
Acceptability & \textit{People judge the store by how fresh and high-quality the food is. Cleanliness of the store and how staff treat customers strongly affect whether people feel good about shopping here. When the store feels dirty, the service is rude, or the food does not meet personal standards, shoppers see it as unacceptable.} \\
\bottomrule
\end{tabular}
\end{table*}

We converted similarities to probabilities using a temperature-scaled softmax,
\begin{equation}
P(j\mid k)=\frac{\exp(s_{jk}/\tau)}{\sum_{j'=1}^{5}\exp(s_{j'k}/\tau)},
\label{eq:softmax}
\end{equation}
where the temperature $\tau=0.117$ was selected by minimizing cross-entropy against manually validated topic--dimension assignments. The highest-probability dimension was assigned to every topic. To retain overlapping concepts, we also assigned a second or third dimension when the probability difference from the preceding label was less than $\delta=0.17$. Six topics received more than one dimension label. Each review inherited all labels assigned to its topic.

One researcher manually labeled the retained topics using their keywords and representative reviews. Automated and manual primary labels agreed for 85.4\% of topics; macro- and prevalence-weighted one-versus-rest ROC--AUC values were 0.915 and 0.861, respectively.

\subsection*{Store-level dimension scores}
We combined topic assignments with star ratings to construct a review-level signal. One-star reviews were assigned $g_n=-1$, reviews with two to four stars were assigned $g_n=0$ and five-star reviews were assigned $g_n=+1$. Intermediate ratings, therefore, contributed to the denominator of a store's mean without being treated as unambiguously positive or negative. Reviews assigned to multiple dimensions contributed once to each corresponding store--dimension combination.

For store $\ell$ and dimension $j$, let $I_{\ell j}$ denote the set of reviews assigned to that store and dimension. The empirical store--dimension score was
\begin{equation}
\bar{s}_{\ell j}=\frac{1}{|I_{\ell j}|}\sum_{n\in I_{\ell j}}g_n.
\label{eq:store_mean}
\end{equation}

\subsection*{Bayesian partial pooling}
Review volume differed across stores and dimensions. We therefore estimated store scores using a separate Bayesian hierarchical model for each dimension. The observed mean was modeled as
\begin{align}
\bar{s}_{\ell j}\mid\theta_{\ell j}&\sim\mathcal{N}(\theta_{\ell j},\mathrm{SE}_{\ell j}^{2}),\\
\theta_{\ell j}\mid\mu_j,\sigma_j&\sim\mathcal{N}(\mu_j,\sigma_j^{2}),
\end{align}
with weakly informative priors $\mu_j\sim\mathcal{N}(0,0.5^2)$ and $\sigma_j\sim\mathrm{HalfNormal}(0.5)$. The sampling standard error was derived from the empirical proportions of positive and negative review signals. Posterior inference used the No-U-Turn Sampler implemented in PyMC \cite{carpenter_hierarchical_2016,dovgalecs_seybolt_luhmann_2021}. We used the posterior mean $\hat{\theta}_{\ell j}$ as the partially pooled estimate and standardized estimates within each dimension across stores. Thus, a score of zero denotes the regional mean for that dimension and a score of one denotes one standard deviation above that mean.

\subsection*{Census-tract exposure scores}
We translated store-level scores to census tracts using a distance-decay spatial interaction model \cite{anderson_theoretical_1979}. For tract $i$ and store $\ell$, the weight was
\begin{equation}
w_{i\ell}=\exp(-D_{i\ell}/h),
\end{equation}
where $D_{i\ell}$ is the distance between the tract centroid and store and $h=1$ mile. The exposure score for tract $i$ and dimension $j$ was the normalized weighted mean
\begin{equation}
E_{ij}=\frac{\sum_{\ell}w_{i\ell}z_{\ell j}}{\sum_{\ell}w_{i\ell}},
\label{eq:tract_exposure}
\end{equation}
where $z_{\ell j}$ is the standardized partially pooled store score.

\subsection*{Statistical analysis}
We measured pairwise associations among store-level dimension scores using Pearson correlations with pairwise-complete observations. To examine socioeconomic patterning, we estimated population-weighted least-squares models at the census-tract level. Outcomes were distance to the nearest grocery store or $E_{ij}$. Covariates were centred median household income, centred proportion of residents identifying as Black and their interaction. Tract population was used as the regression weight. All tests were two-sided, and statistical significance was evaluated at $P<0.05$.

\bibliography{bibcao3mar,manual}

\clearpage
\section*{Supplement}

\subsection*{Store and Dimension Distribution}

\begin{table*}[!h]
\centering
\caption{Distribution of Raleigh grocery-store reviews and food-access dimensions across retail chains. Values in the five dimension columns are the number of reviews assigned to that dimension, with the percentage of all reviews for that chain in parentheses. Reviews could be assigned to more than one dimension; consequently, dimension percentages within a row need not sum to 100\%. The percentage beside each chain's review count is its share of the full sample of 25,125 reviews.}
\label{tab:chain_review_distribution}
\scriptsize
\setlength{\tabcolsep}{3pt}
\begin{tabular*}{\textwidth}{@{\extracolsep{\fill}}lrrrrrrr@{}}
\toprule
Retail chain & Stores, $n$ & Reviews, $n$ (\%) & Availability, $n$ (\%) & Accessibility, $n$ (\%) & Accommodation, $n$ (\%) & Affordability, $n$ (\%) & Acceptability, $n$ (\%) \\
\midrule
Harris Teeter          & 14 & 7,131 (28.4\%) & 158 (2.2\%) & 411 (5.8\%) & 245 (3.4\%) & 656 (9.2\%)  & 3,833 (53.8\%) \\
Food Lion              & 16 & 5,096 (20.3\%) & 134 (2.6\%) & 319 (6.3\%) & 202 (4.0\%) & 502 (9.9\%)  & 2,696 (52.9\%) \\
Wegmans                &  2 & 3,258 (13.0\%) &  89 (2.7\%) &  72 (2.2\%) &  38 (1.2\%) & 304 (9.3\%)  & 1,287 (39.5\%) \\
ALDI                    &  4 & 2,214 (8.8\%)  &  66 (3.0\%) &  91 (4.1\%) &  64 (2.9\%) & 576 (26.0\%) &   907 (41.0\%) \\
Publix                  &  3 & 1,660 (6.6\%)  &  30 (1.8\%) &  97 (5.8\%) &  16 (1.0\%) & 144 (8.7\%)  &   974 (58.7\%) \\
Lidl                    &  3 & 1,550 (6.2\%)  &  39 (2.5\%) &  44 (2.8\%) &  41 (2.6\%) & 304 (19.6\%) &   658 (42.5\%) \\
Whole Foods Market      &  2 & 1,327 (5.3\%)  &  43 (3.2\%) &  41 (3.1\%) &  22 (1.7\%) & 165 (12.4\%) &   679 (51.2\%) \\
Lowes Foods             &  2 & 1,281 (5.1\%)  &  23 (1.8\%) &  31 (2.4\%) &  30 (2.3\%) & 146 (11.4\%) &   749 (58.5\%) \\
Carlie C's IGA          &  1 &   714 (2.8\%)  &  18 (2.5\%) &  15 (2.1\%) &   9 (1.3\%) & 115 (16.1\%) &   388 (54.3\%) \\
Sprouts Farmers Market  &  1 &   626 (2.5\%)  &  38 (6.1\%) &  12 (1.9\%) &   4 (0.6\%) &  87 (13.9\%) &   362 (57.8\%) \\
The Fresh Market        &  1 &   268 (1.1\%)  &   8 (3.0\%) &   5 (1.9\%) &   2 (0.7\%) &  33 (12.3\%) &   135 (50.4\%) \\
\midrule
Total                   & 49 & 25,125 (100.0\%) & 646 (2.6\%) & 1,138 (4.5\%) & 673 (2.7\%) & 3,032 (12.1\%) & 12,668 (50.4\%) \\
\bottomrule
\end{tabular*}
\end{table*}

\subsection*{Single and Multi-Reviewer Comparison}

\begin{table*}[!h]
\centering
\caption{Food-access dimensions identified in reviews written by people who reviewed one versus multiple stores in the Raleigh sample. Reviewer type is defined by the number of distinct stores reviewed within the 49-store study sample. Values in the five dimension columns are the number of reviews assigned to that dimension, with the percentage of all reviews in that reviewer group in parentheses. Reviews could be assigned to more than one dimension; consequently, dimension percentages within a row need not sum to 100\%. The difference row reports multiple-store minus one-store reviewers in percentage points. The descriptive association between reviewer group and dimension assignment was small (Cramer's $V=0.05$).}
\label{tab:reviewer_scope_dimension_distribution}
\scriptsize
\setlength{\tabcolsep}{3pt}
\resizebox{\textwidth}{!}{%
\begin{tabular}{lrrrrrrr}
\toprule
Reviewer group & Reviewers, $n$ & Reviews, $n$ (\%) & Availability, $n$ (\%) & Accessibility, $n$ (\%) & Accommodation, $n$ (\%) & Affordability, $n$ (\%) & Acceptability, $n$ (\%) \\
\midrule
One store       & 13,878 & 13,878 (55.2\%) & 343 (2.5\%) & 645 (4.6\%) & 409 (2.9\%) & 1,538 (11.1\%) & 7,194 (51.8\%) \\
Multiple stores &  3,852 & 11,247 (44.8\%) & 303 (2.7\%) & 493 (4.4\%) & 264 (2.3\%) & 1,494 (13.3\%) & 5,474 (48.7\%) \\
\addlinespace
Difference, percentage points & --- & --- & +0.2 & $-0.3$ & $-0.6$ & +2.2 & $-3.2$ \\
\midrule
Total           & 17,730 & 25,125 (100.0\%) & 646 (2.6\%) & 1,138 (4.5\%) & 673 (2.7\%) & 3,032 (12.1\%) & 12,668 (50.4\%) \\
\bottomrule
\end{tabular}%
}
\end{table*}
\begin{table*}[!h]
\centering
\caption{Google star ratings and review-level food-access scores by reviewer scope and dimension in the Raleigh sample. Multiple reviewers are individuals who reviewed more than one store in our dataset. Values are means, with standard deviations in parentheses. Google stars range from 1 to 5. Following the store-scoring procedure, one-star reviews receive a review score of $-1$, two- to four-star reviews receive 0, and five-star reviews receive $+1$; all ratings remain in the denominator. Dimension-specific values include reviews assigned to the corresponding dimension, and reviews assigned to multiple dimensions contribute to each applicable column.}
\label{tab:reviewer_scope_dimension_scores}
\small
\setlength{\tabcolsep}{5pt}
\resizebox{\textwidth}{!}{%
\begin{tabular}{llrrrrrr}
\toprule
Reviewer group & Measure & Overall & Availability & Accessibility & Accommodation & Affordability & Acceptability \\
\midrule
One store       & Google stars & 4.23 (1.26) & 4.31 (1.10) & 3.35 (1.52) & 2.76 (1.55) & 4.31 (1.05) & 4.07 (1.41) \\
Multiple stores & Google stars & 4.36 (1.01) & 4.34 (0.94) & 3.68 (1.31) & 3.32 (1.49) & 4.38 (0.90) & 4.32 (1.08) \\
\addlinespace
One store       & Review score & 0.545 (0.652) & 0.571 (0.582) & 0.118 (0.717) & $-0.130$ (0.721) & 0.554 (0.575) & 0.473 (0.707) \\
Multiple stores & Review score & 0.574 (0.562) & 0.531 (0.550) & 0.252 (0.619) & 0.117 (0.690) & 0.558 (0.538) & 0.560 (0.584) \\
\bottomrule
\end{tabular}%
}
\end{table*}

\end{document}